\documentclass[
 twocolumn,
]{ceurart}

\usepackage{listings}
\usepackage{wrapfig} 
\usepackage{placeins} 

\begin{document}

\copyrightyear{2026}
\copyrightclause{Copyright for this paper by its authors.
  Use permitted under Creative Commons License Attribution 4.0
  International (CC BY 4.0).}

\conference{RecSys in HR '26: The 6th Workshop on Recommender Systems for Human Resources, in conjunction with the 20th ACM Conference on Recommender Systems, September 28--October 2, 2026, Minneapolis, MN, United States.}

\title{When Occupation-derived Skills Don’t Help: On the Limits of Skill-Augmented Career Path Prediction}

\author[1]{Ege S\"ualp}[%
email=e.sualp@campus.lmu.de,
]
\cormark[1]
\fnmark[1]
\address[1]{MaiNLP, Center for Information and Language Processing, LMU Munich, Germany}

\author[1,2]{Elena Senger}[%
email=elena.senger@cis.lmu.de
]
\cormark[1]
\fnmark[1]
\address[2]{Fraunhofer Institute for Systems and Innovation Research ISI, Germany}

\author[3]{Toine Bogers}[%
email=tobo@itu.dk
]
\address[3]{NERDS, IT University Copenhagen, Denmark}

\author[1]{Barbara Plank}[%
email=b.plank@lmu.de,
]

\cortext[1]{Corresponding author.}
\fntext[1]{These authors contributed equally.}

\begin{abstract}
Skills are widely assumed to play a central role in career transitions, motivating their use in career path prediction (CPP) systems. In this work, we critically examine whether explicit skill representations provide complementary signal beyond text-based job representations. We evaluate multiple strategies for incorporating skills into CPP models, including score fusion and neural architectures, under two conditions: predicted skills and ground-truth ESCO annotations. Surprisingly, across two benchmark datasets, none of these settings yield substantial improvements over strong text-only baselines. Even when using ground-truth skills, gains remain marginal and inconsistent. Through detailed analysis, we identify key factors underlying this result: high redundancy between textual and skill representations, limited discriminative power of ESCO’s taxonomy, and representational collapse in skill embeddings. Furthermore, we find that skill prediction difficulty correlates with overall sample difficulty, limiting the potential for complementary signals. Our findings challenge the common assumption that explicit skill modeling inherently benefits career prediction and suggest that dense text representations already capture most skill-relevant information.
\end{abstract}


\maketitle

\section{Introduction}

Career Path Prediction (CPP), the task of forecasting an individual’s next occupation based on their professional history, has gained increasing attention in both academia and industry \cite{towards_more_realistic_2025}, \cite{gasco2025overviewtalentclef2025skill}. Accurate CPP models enable a wide range of applications, including personalized career guidance, workforce planning, and talent management, to name a few examples. With the growing availability of large-scale career data and advances in natural language processing (NLP), recent approaches model career trajectories as sequences of textual descriptions such as job titles and responsibilities, learning predictive patterns directly from unstructured text.

Despite strong empirical performance, these text-based approaches rely on an implicit assumption: that textual representations sufficiently capture the underlying drivers of career transitions. However, from a labor market perspective, skills—the transferable competencies acquired through experience—are widely regarded as the primary mechanism governing occupational mobility. This has led to a natural hypothesis: explicitly modeling skills should provide additional, structured signal that improves career path prediction beyond text-only baselines.
Recent developments make this hypothesis practically testable. Large-scale taxonomies such as ESCO \footnote{European Skills, Competences, Qualifications and Occupations: \url{https://esco.ec.europa.eu/en/about-esco/what-esco}} provide standardized mappings between occupations and skills, while modern retrieval and language models enable the extraction of structured skill representations from unstructured career histories. Together, these advances suggest a promising direction: augmenting CPP models with explicit skill information inferred from text.
However, obtaining reliable skill representations from unstructured career histories is non-trivial, as the same role can be expressed in highly varied ways across individuals and contexts. We address this through a retrieval-based pipeline that grounds job titles and descriptions in ESCO occupations, from which associated skills are derived. Hierarchical occupational structure is further leveraged to improve skill coverage and consistency.

In this work, we take a step back and critically examine this assumption. Rather than focusing on improving skill extraction, we ask a more fundamental question: do explicit skill representations actually improve career path prediction, and if not, why do they fail to provide complementary signal? To answer this, we conduct a systematic study across multiple skill sources and integration strategies. Specifically, we evaluate two  settings: (1) predicted skills inferred from occupations and (2) ground-truth ESCO skill annotations. This allows us to disentangle the impact of skill quality from the effectiveness of skill integration.
Surprisingly, our results show that \emph{none of these settings yield substantial improvements over strong text-only baselines}. Even when using ground-truth skills, gains remain marginal and inconsistent. This finding challenges the common assumption that explicit skill modeling inherently provides complementary signal for CPP.
To better understand this result, we perform an in-depth analysis and identify three key factors. First, we observe a high degree of redundancy between textual and skill representations, suggesting that modern text embeddings already encode much of the relevant skill information. Second, we find that skill embeddings exhibit limited discriminative power, often collapsing into coarse occupational clusters. Third, we show that improvements from skill-based signals occur primarily in lower-ranked predictions, while degrading performance in the most critical top-k region.

Taken together, our findings suggest that the limited effectiveness of skill-enhanced CPP is not merely due to imperfect skill extraction, but reflects a more fundamental issue in how skills are represented and utilized in current models. This has important implications for the design of skill-aware recommender systems and calls for a re-evaluation of how structured knowledge is integrated with dense representations.

\section{Related Work}


\subsection{Career Path Prediction}

Career path prediction has evolved from structured sequence modeling toward text-based representation learning. Earlier work uses structured representations such as occupation graphs, skill graphs, factorization models, and recurrent sequence models to capture transitions between jobs~\citep{Groot2021JobPK,Gugnani.2018, li_et_al_2017,CerillaSVF23,ahead_zhang,schellingerhout2022explainable}. More recent work has shifted toward textual occupation representations and retrieval-based formulations of CPP \citep{decorte2023careerpathpredictionusing, towards_more_realistic_2025}. 

Skills are widely regarded as a natural bridge between candidate profiles and occupational requirements, motivating their integration into career and job recommendation systems \cite{decorte2023careerpathpredictionusing,guan2024jobformerskillawarejobrecommendation}. Prior work shows that career trajectories encode latent preferences beyond current qualifications \citep{gong2024career}, while skill-aware representation learning improves the modeling of occupational similarity and person-job fit \citep{guan2024jobformerskillawarejobrecommendation,gugnani2020implicit}. Furthermore, \citet{zadykian2025explainablejobtitlematching} demonstrate that exploiting the ESCO skill hierarchy through graph neural networks improves job-title matching, suggesting that taxonomic skill information provides complementary signal beyond text embeddings. Most closely related to our work, \citet{decorte2023careerpathpredictionusing} combine text-based career representations with ESCO skill-overlap features and report improvements over text-only models. However, their evaluation assumes pre-mapped ESCO occupations rather than free-text career histories. Building on the text-only system of \citet{towards_more_realistic_2025}, we examine whether skill integration yields consistent gains under more realistic grounding conditions across two benchmark datasets.

\subsection{Skill Extraction and Grounding}

Mapping free-text job information to standardized taxonomies is a core challenge in labor market analysis. Rather than extracting explicit skill mentions from text, grounding-based approaches retrieve the most semantically similar ESCO occupations or skills for a given job title and description, leveraging dense retrieval models fine-tuned on occupation-skill corpora. TalentCLEF 2025 \cite{gasco2025overviewtalentclef2025skill} provides a shared task benchmark covering both job-to-occupation and job-to-skill mapping, with top-performing systems demonstrating that retrieval-based grounding achieves competitive results on ESCO standardization. However, grounding quality remains imperfect — even state-of-the-art systems score below 0.5 MAP on skill matching \cite{vachharajani2025pjmathematician} — which directly limits the quality of skills available for downstream tasks such as CPP. Our pipeline adopts this grounding-based paradigm, combining job-to-occupation and job-to-skill retrieval with hierarchical fusion over ESCO's taxonomic structure.

\section{Methodology}
We treat CPP as an occupation-ranking task over ESCO. Given a time-ordered career history, where each experience contains a job title and optionally a description, the model ranks candidate ESCO occupations according to their likelihood of being the next occupation. We treat skill inference as a retrieval problem over ESCO skills: for each job experience, the system ranks skills that are likely to be relevant. ESCO distinguishes between essential and optional skills; we consider both as positive skill evidence. Skill-enhanced CPP then evaluates whether adding predicted or oracle skills to text-based career representations improves the final occupation ranking. The following paragraphs describe the methods used to test this.

\paragraph{Skill Prediction Pipeline}
\label{method:skill-prediction-pipeline}




We adopt a multi-step retrieval pipeline grounded in the ESCO taxonomy, building on the job-to-skill and job-to-occupation retrieval tasks of TalentCLEF 2025~\cite{gasco2025overviewtalentclef2025skill}. We first retrieve candidate ESCO skills for each job experience via dense semantic retrieval (job-to-skill matching). We then apply occupation-mediated filtering: job experiences are mapped to their most similar ESCO occupations, and the resulting occupation-skill associations constrain the candidate pool to domain-relevant skills. Finally, ISCO codes are predicted from free-text job representations using a trained neural classifier, and the predicted ISCO group distribution is used to reweight the filtered skill candidates, exploiting the hierarchical structure of the taxonomy. This configuration was selected after comparing it against simpler alternatives: direct retrieval alone substantially underperforms (MAP 0.157), occupation-mediated filtering alone already recovers most of the gain (MAP 0.408–0.428 depending on the grounding encoder), and ISCO-weighted reweighting adds a further, smaller improvement (MAP up to 0.452). Building on this pipeline, we evaluate multiple skill prediction variants to analyze how skill prediction quality affects downstream CPP performance.

\paragraph{Skill-Enhanced Career Path Prediction}

We build on the strong text-only MLP of \citet{towards_more_realistic_2025} as our baseline, which encodes career histories via a sentence transformer and ranks candidate ESCO occupations by cosine similarity. We investigate several strategies for incorporating predicted skill information into this framework, ranging from score fusion to neural architectures.

\paragraph{Skill Overlap and Fusion}

Following \citet{decorte2023careerpathpredictionusing}, we implement a skill overlap baseline that measures the fraction of a target occupation's ESCO-defined skills covered by the predicted skills accumulated across the candidate's career history. This provides an explicit skill alignment signal independent of text-based representations.
We combine this signal with the text-based model score using two fusion strategies: linear interpolation (similar to ~\citet{decorte2023careerpathpredictionusing}) and multiplicative reweighting. Fusion hyperparameters are selected via grid search on the validation set, optimising for MRR.

\paragraph{Skill-Enhanced Neural Models}
\label{method:skill-aware-neural-models}
We extend the strong text-only MLP baseline with skill-aware variants, exploring two orthogonal design dimensions: how skills are encoded, and how they are fused with career text representations.

Regarding how skills are encoded, we evaluate two alternative approaches to encode skills into a single representation. \emph{Skill pooling} encodes each predicted skill independently via a pretrained sentence transformer and aggregates them into a single embedding via weighted pooling across the career history, where weights reflect retrieval confidence, skill specificity (IDF scores), and job recency (logarithmic job-order decay). \emph{Skill concatenation} instead joins the top-$k$ predicted skills into a single string and encodes it in one pass, allowing the encoder to capture inter-skill relationships within context.

Both strategies are combined with the career text embedding using either early fusion, concatenating both embeddings into a single MLP input, or late fusion, where separate MLP stacks process each modality independently before a shared projection layer. All models are trained using cosine similarity against target occupation embeddings.

\section{Experimental Setup}

\paragraph{Datasets and Taxonomy}

We evaluate our approach on two publicly available datasets linked to the ESCO taxonomy: DECORTE \cite{decorte2023careerpathpredictionusing} and KARRIEREWEGE+ \cite{towards_more_realistic_2025}. Both datasets provide career histories mapped to standardized ESCO occupation labels. 
We primarily report results on DECORTE, as it includes human-written free-text job descriptions and therefore better reflects real-world resume data. The dataset contains 2,164 career histories with 9,885 job experiences mapped to 1,155 unique ESCO occupations. We use the official train/validation/test splits and follow prior work by augmenting training data with sub-sequences of career histories. KARRIEREWEGE+ provides a complementary larger-scale dataset built from anonymized German employment records; career trajectories are real, but job titles and descriptions are synthetically generated to provide free-text input. It comprises 99,997 career histories with 502,641 job experiences mapped to 1,189 unique ESCO occupations, used to assess generalisability.

The ESCO taxonomy serves both as the label space for occupation prediction and as the source of skill information. It contains over 3,000 occupations and nearly 14,000 skills, linked through essential and optional relations. Following prior work, we treat both relation types as positive signals.

\paragraph{Evaluation Metrics}
We evaluate CPP with Mean Reciprocal Rank (MRR) and Recall@$k$, standard ranking metrics that capture top-rank precision and whether the target appears among the top-$k$ recommendations. Skill prediction is evaluated with Mean Average Precision (MAP), which captures ranking quality across the full set of relevant skills.


\paragraph{Career Path Prediction Setup} In the text-based framework of \citet{towards_more_realistic_2025}, each job experience is formatted as a concatenation of title and description, encoded using \texttt{MPNet-Decorte},\footnote{\url{https://huggingface.co/ElenaSenger/career-path-representation-mpnet-decorte}} a sentence transformer \cite{reimers2019sentencebertsentenceembeddingsusing} contrastively trained on free-text career data. An MLP maps career history embeddings to occupation embeddings; candidate occupations are ranked at inference by cosine similarity against precomputed ESCO occupation embeddings. For all runs, we use batch size 16, learning rate $2{\times}10^{-5}$, hidden dimension 512, single hidden layer, and dropout 0.1, trained with Cosine Embedding Loss and early stopping (patience 2) on validation loss; this configuration follows \citet{towards_more_realistic_2025}, as deeper networks consistently underperformed in preliminary experiments. Training data is augmented with contiguous career path sub-sequences of minimum length 2. We evaluate on both the augmented test split ($N{=}1{,}802$) and the original clean split ($N{=}227$). 

\paragraph{Skill Prediction and Integration} We evaluate our skill extraction pipeline (Section~\ref{method:skill-prediction-pipeline}) on DECORTE using MAP, comparing direct job-to-skill retrieval against ISCO-weighted fusion variants. For downstream CPP experiments, we use the JobBERT-based ISCO-weighted fusion (H2, MAP 0.435), which yields better discriminative signal for occupation ranking than the highest-MAP configuration (JobGTE + H2, MAP 0.452). To disentangle skill quality from integration strategy, we additionally evaluate an oracle condition using skills derived from gold ESCO occupation labels, applied to the score fusion setting. We evaluate two integration strategies: (i)~\textbf{Score fusion}, combining text-based and skill overlap scores with hyperparameters selected by grid search optimising for MRR; and (ii)~\textbf{Skill-enhanced neural models}, extending the MLP with skill embeddings as described in Section~\ref{method:skill-aware-neural-models}, with the best configuration further tuned using Optuna.

\section{Results}

\subsection{Text-Only Baseline}

On DECORTE, we first reproduce the text-only baseline from \cite{towards_more_realistic_2025}, achieving comparable performance (MRR 0.266 vs.\ 0.256 reported); corresponding results on KARRIEREWEGE+ follow in Section~\ref{subsec:scalability}.
Including job descriptions improves performance over titles alone (MRR 0.266 vs.\ 0.224), highlighting the importance of richer textual context. 
Results on the clean test split are slightly higher in MRR (0.276 vs.\ 0.266) and R@5 (0.374 vs.\ 0.356), but show lower recall at larger cutoff (R@10 0.414 vs.\ 0.420, from Table~\ref{tab:neural-results}).

\subsection{Skill Overlap Scoring}

We evaluate skill overlap both as a standalone signal and in combination with text-based predictions.
Using ground-truth ESCO skills as an oracle yields substantially lower performance than the text-only model (MRR 0.198 vs.\ 0.266). Using predicted skills further reduces performance, with direct retrieval performing poorly due to noise.

Applying filtering and taxonomic fusion improves skill-based scoring, with the best configuration approaching oracle performance. However, all variants remain below the text-only baseline. When combined with text-based predictions, skill overlap provides no measurable improvement. Fusion methods consistently assign negligible weight to the skill component.

\subsection{Skill-Enhanced Neural Models}
We evaluate MLP models augmented with skill representations, varying both the skill integration method (pooling vs.\ concatenation) and the fusion architecture (early vs.\ late), as described in Section~\ref{method:skill-aware-neural-models}. Table~\ref{tab:neural-results} reports results for the best configuration of each integration method.

Overall, skill-enhanced models achieve only marginal improvements over the text-only baseline. On the augmented test split, differences are negligible (MRR 0.270 vs.\ 0.266). On the clean split, gains are somewhat larger (up to MRR 0.287 with log-weighted pooling), but correspond to only a handful of additional correct predictions given the smaller sample size ($N{=}227$).

Comparing integration methods, skill pooling consistently outperforms concatenation on MRR (0.270 vs.\ 0.262 augmented; 0.283 vs.\ 0.278 clean), while concatenation achieves the better R@10 (0.449 vs.\ 0.427 clean). This suggests skills can occasionally help recover the correct occupation at lower ranks, but do not sharpen top-1 predictions, consistent with the redundancy we examine in Section~\ref{sec:discussion}.

Fusion architecture (early vs.\ late) and encoder choice have comparatively little effect: differences between configurations within each integration method are within 0.01 MRR, smaller than the gap between integration methods themselves. Task-specific skill encoders ({\tt pjmath-BGE}) modestly outperform general-purpose alternatives, but all configurations remain close to the text-only baseline.

Across configurations, results are largely stable:
\begin{itemize}
    \item Different encoders yield similar performance, with slight advantages for task-specific models.
    \item Early and late fusion architectures perform comparably.
    \item Pooling and weighting strategies have minimal impact.
\end{itemize}


\begin{table*}[t]
\centering
\caption{Skill-enhanced MLP results, by skill integration method. For each method, the best-performing configuration is reported.}
\label{tab:neural-results}
\begin{tabular}{llcccccc}
\toprule
& & \multicolumn{3}{c}{Augmented} & \multicolumn{3}{c}{Clean} \\
Dataset & Configuration & MRR & R@5 & R@10 & MRR & R@5 & R@10 \\
\midrule
\multirow{3}{*}{DECORTE}
& Text-only baseline       & 0.266 & \textbf{0.356} & 0.420 & 0.276 & 0.374 & 0.414 \\
& Skill Pooling             & \textbf{0.270} & \textbf{0.356} &\textbf{0.426} & \textbf{0.283} & \textbf{0.396} & 0.427 \\
& Concatenated Encoding     & 0.262 & 0.342 & 0.425 & 0.278 & 0.392 & \textbf{0.449} \\
\midrule
\multirow{3}{*}{KARRIEREWEGE+}
& Text-only baseline       & 0.439 & 0.527 & 0.604 & 0.334 & 0.425 & 0.502 \\
& Skill Pooling             & 0.438 & 0.525 & 0.602 & 0.334 & 0.421 & 0.502 \\
& Concatenated Encoding     & \textbf{0.441} & \textbf{0.529} & \textbf{0.607} & \textbf{0.338} & \textbf{0.427} & \textbf{0.508} \\
\bottomrule
\end{tabular}
\end{table*}

\subsection{Scalability on Larger Data}
\label{subsec:scalability}
To assess whether our findings generalize beyond DECORTE, we extend evaluation to KARRIEREWEGE+, which is $\sim$40$\times$ larger ($N{\approx}100{,}000$ career histories) but uses LLM-paraphrased synthetic descriptions rather than free text. Since dataset size and text type differ simultaneously, this comparison indicates whether the pattern holds at scale rather than isolating the effect of scale itself.

Absolute performance is substantially higher than on DECORTE (text-only MRR 0.439 vs.\ 0.266), likely reflecting easier career trajectories in KARRIEREWEGE+, where transitions remain within the same occupation more often than in DECORTE \citep{towards_more_realistic_2025}. Skill enhancement again yields only marginal gains: concatenation improves MRR by 0.5\% (0.441 vs.\ 0.439) on the augmented split and 1.2\% (0.508 vs.\ 0.502) on clean, while pooling shows no improvement. Both effects are smaller than on DECORTE (+1.5\% and +4.0\% respectively), suggesting that skill redundancy is, if anything, more pronounced with synthetic text. As on DECORTE, hybrid fusion of text and skill-overlap scores converges to fully ignoring the skill component ($\alpha=1.0$), confirming that skill-based scoring provides no complementary signal at this scale either.

\section{Discussion}
\label{sec:discussion}

This section examines why skill integration yields limited gains, analyzing prediction scores, embedding representations, and a structural limitation in skill attribution. Across these analyses, a consistent picture emerges: the limited effect of skill integration is not an artifact of model architecture or skill prediction quality, but reflects a deeper redundancy between textual and skill-based signals.

\subsection{Signal Redundancy}
\label{subsec:redundancy}

To understand why skill enhancement fails to improve aggregate performance, we examine the relationship between text-based and skill-based prediction signals. For each test sample, we compute the Spearman correlation \cite{spearman1904proof} between the ranking scores produced by the text-only and skill-only models across all candidate occupations. The mean correlation is 0.79 (median 0.84), indicating that the two modalities share roughly 80\% of their ranking signal. This redundancy explains why explicitly adding skill information provides limited complementary value: contrastively-trained text encoders already capture much of what predicted skills offer.

Contingency analysis of top-1 correctness reinforces this picture. The two models agree in 90.9\% of cases (both correct: 12.4\%; both incorrect: 78.5\%); text-only corrects the skill-only model's errors in 6.2\% of samples, while the reverse occurs in only 2.9\%. This 2.9\% represents the maximum achievable improvement from skill integration under any fusion strategy, and is consistent with the marginal gains observed. Where models disagree, text-only is also more reliable (10.5\% vs.\ 5.0\% accuracy on the disagreement set), so a hybrid model correctly defers to text-based signal in most conflict cases. This deference extends to candidate selection itself: across \(k \in \{5, 10, 20\}\), the hybrid model's candidate sets consistently align more closely with text-only (63.0–66.5\% Jaccard overlap) than with skill-only (45.8–51.0\%), confirming that score fusion draws its predictions predominantly from the text modality regardless of cutoff.

Redundancy is not uniform across all cases, however. Stratifying by career trajectory type (Table~\ref{tab:trajectory-effects}) shows that skill enhancement provides its largest benefit for regressions, i.e., transitions from managerial to non-managerial roles ($+0.020$ MRR)---the only trajectory type with a clear positive effect, plausibly because shifting skill profiles signal downward mobility that text alone underrepresents. This benefit is confined to a small subset (13.8\% of samples), however, and is offset by slight degradation on the dominant lateral transitions (78.6\%), as well as on occupations with distinctive skill profiles more broadly, such as Cooks (+0.185 MRR).

\subsection{Discriminative Power and Confident-Prediction Disruption}
\label{subsec:calibration}

\begin{figure*}
    \centering
    \includegraphics[width=1.05\textwidth]{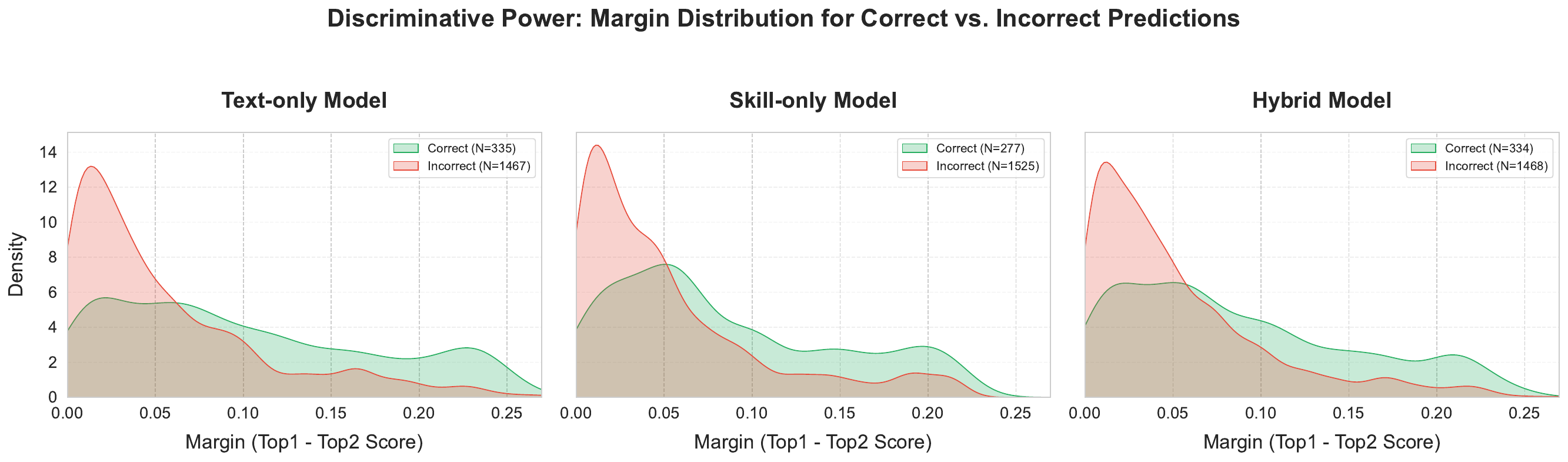}
    \caption{Discriminative power analysis via score margins ($M = s_{top-1} - s_{top-2}$). Distributions show the density of correct predictions (green) versus incorrect ones (red) for Text, Skill, and Hybrid models. A higher margin for correct predictions indicates superior calibration and decision certainty. The sharper separation in the Text-only model signifies higher discriminative power, whereas the overlap in the Skill-only model suggests lower certainty in distinguishing between occupational classes.}
\label{fig:discriminative-power}
\end{figure*}

Beyond aggregate redundancy, we examine whether skill-only predictions are well-calibrated, i.e., whether the margin between the top- and second-ranked candidate reliably signals correctness. Skill-only trails text-only on accuracy (0.154 vs.\ 0.186), mean margin (0.057 vs.\ 0.065), and calibration correlation between margin and correctness (0.213 vs.\ 0.276). As Figure~\ref{fig:discriminative-power} shows, skill-only margins concentrate narrowly around 0.05 regardless of outcome, while text-only exhibits a flatter distribution with clearer separation between correct and incorrect predictions.

This has a concrete effect on performance: skill enhancement tends to hurt cases where the text encoder is already confident. When the target occupation matches the most recent one (same-role continuation, $N{=}290$), text-only alone already reaches MRR 0.713, yet skill enhancement \emph{lowers} performance here ($-0.013$ MRR, $-1.4\%$ R@1). On the harder and far more common role-change scenario ($N{=}1{,}512$), the benefit is marginal ($+0.002$ MRR). Because straightforward cases make up most of the test set, this pattern anticipates the rank-movement results below: skill enhancement disrupts predictions the text model is already getting right.

\subsection{The Rank Movement Paradox}
\label{subsec:rank-movement}

\begin{figure}
\centering
\includegraphics[width=\columnwidth]{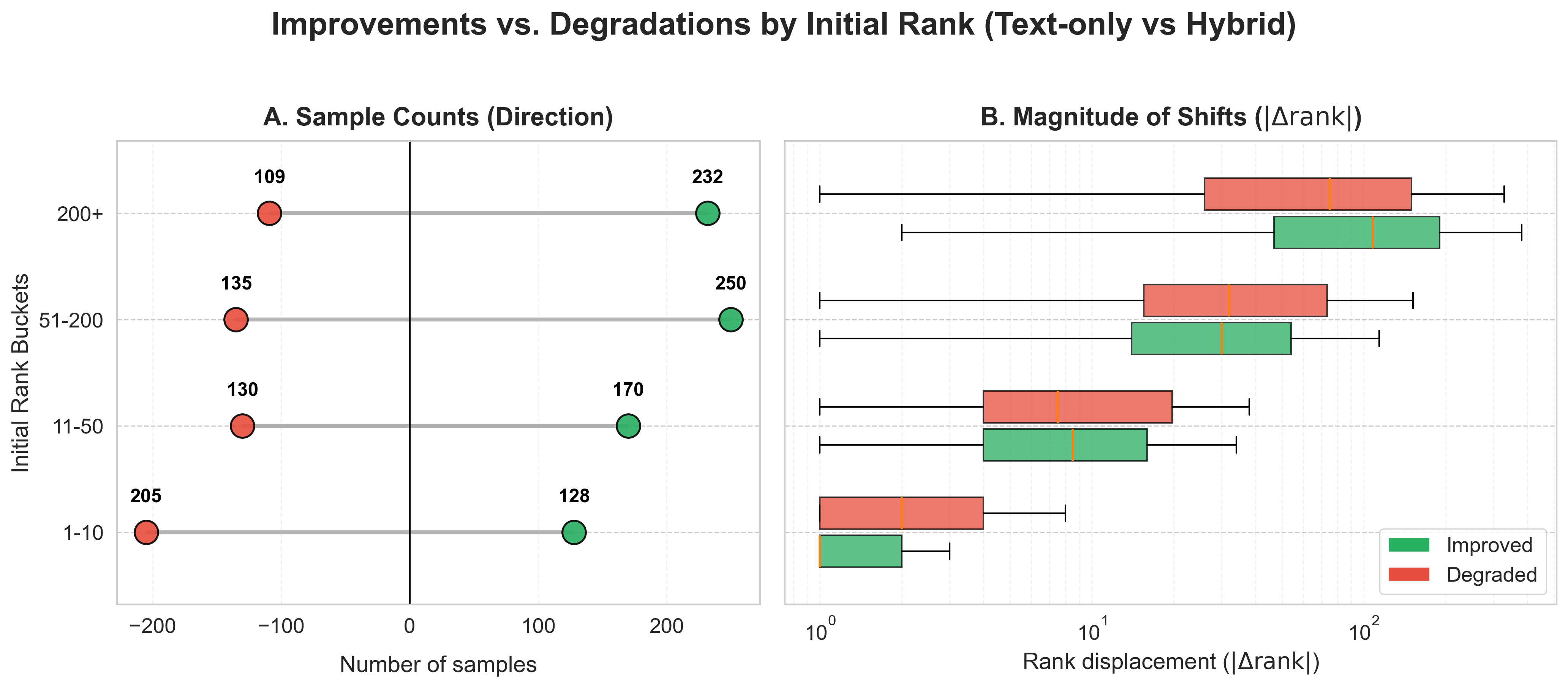}
\caption{Rank dynamics between text-only and hybrid models. (A) Frequency of improved vs.\ degraded samples by initial rank bucket. (B) Distribution of rank displacement magnitude. Degradations concentrate in high-rank buckets (1--10) but are individually small; improvements occur mainly in low-rank buckets and involve larger shifts that are metric-insensitive.}
\label{fig:rank-movement}
\end{figure}

This disruption pattern is visible at the rank level as well: tracking how skill integration moves the rank of the correct occupation reveals where these effects occur. For each test sample, we compare the rank of the target occupation under the text-only and hybrid models (Figure~\ref{fig:rank-movement}). More samples improve (43.3\%, average +55.4 positions) than degrade (32.1\%, average $-39.1$ positions), which would suggest a net benefit.

The two types of movement, however, occur in different regions of the ranking, so this apparent advantage does not translate into metric gains. Improvements typically lift the target from very low ranks (e.g., position 500 to 450), which affects neither MRR nor Recall@$k$. Degradations, by contrast, cluster near the top of the ranking (e.g., position 8 to 12, or 1 to 3), directly harming both metrics. Skill enhancement therefore tends to help in metric-insensitive regions while occasionally disrupting confident, high-ranked predictions---explaining the paradox of more rank improvements than degradations without any corresponding aggregate gain.

\subsection{Representational Collapse and Target Misalignment}
\label{subsec:collapse}


\begin{table}
\caption{PCA cumulative explained variance comparison between text and skill embeddings.}
\label{tab:pca-variance}
\begin{tabular}{lccc}
\toprule
Embedding & Top 10 & Top 20 & Top 50 \\
\midrule
Text & 38.6\% & 53.7\% & 73.7\% \\
\midrule
\multicolumn{4}{l}{\textit{Pooled Skills (pjmath-BGE)}} \\
Mean pooling   & 80.0\% & 91.6\% & 98.1\% \\
Weighted IDF   & 77.3\% & 90.0\% & 97.7\% \\
Log pooling    & 76.2\% & 89.1\% & 97.3\% \\
\midrule
\multicolumn{4}{l}{\textit{Concatenated Skills (MPNet-Decorte)}} \\
Top-10 weighted & 54.4\% & 68.6\% & 84.8\% \\
\bottomrule
\end{tabular}
\end{table}

\begin{figure*}[p]
\centering
\includegraphics[width=0.85\textwidth]{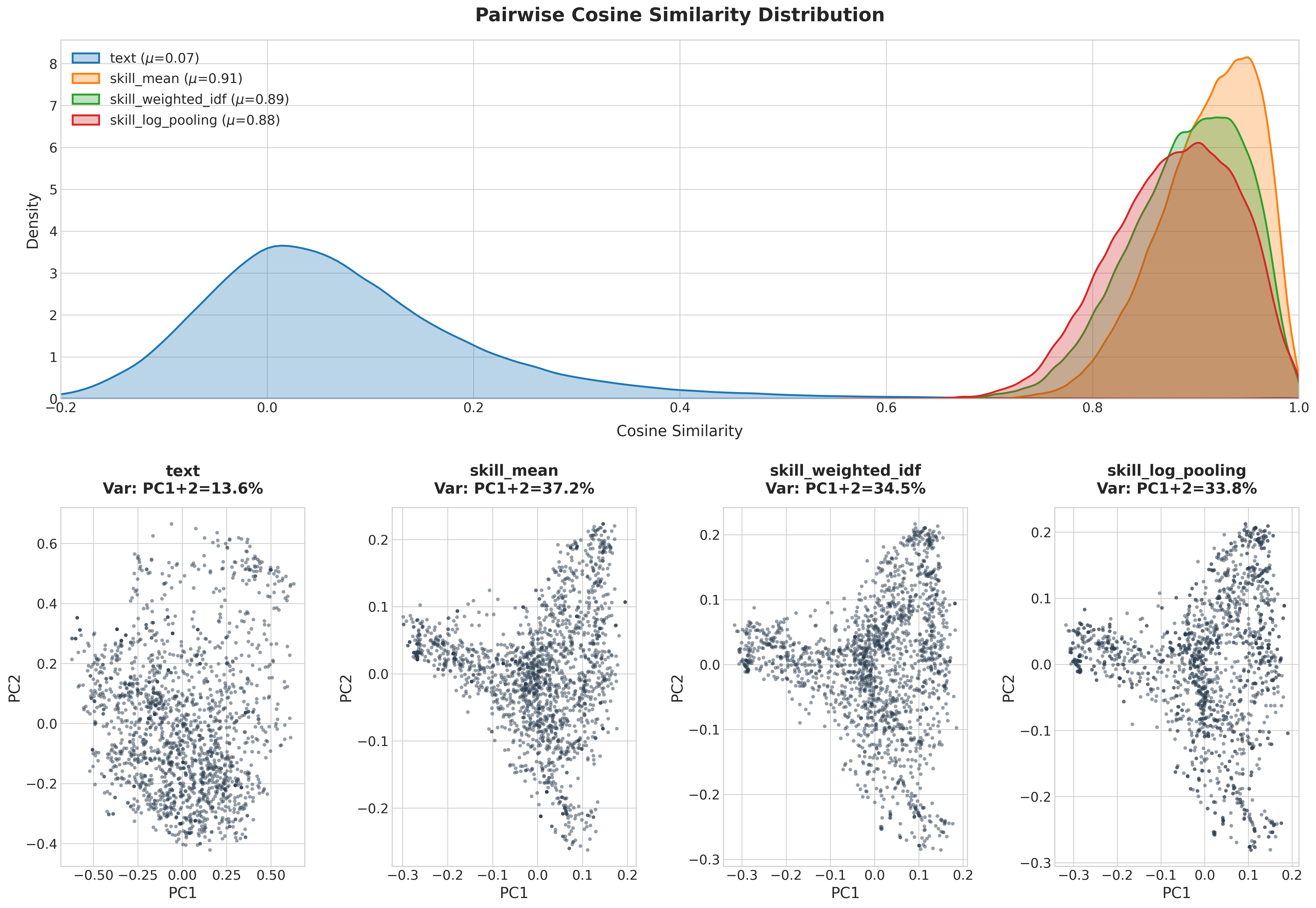}
\caption{Pairwise cosine similarity and PCA projections for text vs.\ pooled skill embeddings. Pooled skill embeddings ($\mu \approx 0.90$) shift dramatically toward near-identical representations compared to text ($\mu = 0.07$); PCA projections show text embeddings spanning the latent space broadly while skill embeddings collapse into constrained, structured clusters.}
\label{fig:pca-pooled}
\end{figure*}

\begin{figure*}[p]
\centering
\includegraphics[width=0.85\textwidth]{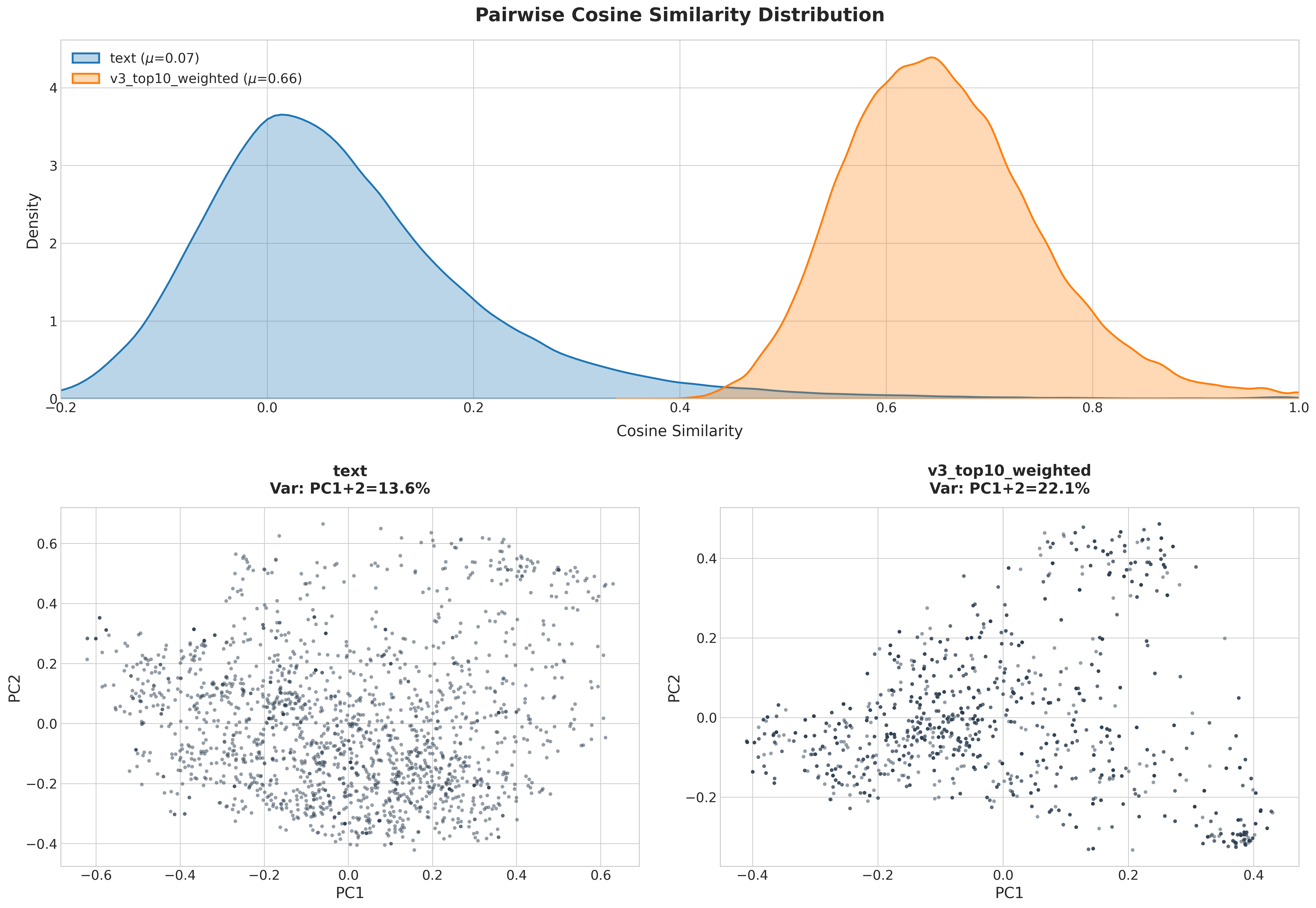}
\caption{Pairwise cosine similarity and PCA projections for text vs.\ concatenated skill embeddings. Even after mitigating collapse via IDF-weighted concatenation ($\mu = 0.66$), the skill modality remains more concentrated than text ($\mu = 0.07$), lacking the same latent diversity.}
\label{fig:pca-concat}
\end{figure*}

A complementary explanation for skill enhancement's limited effect lies in the geometry of the skill embeddings themselves. Text embeddings spread variance broadly across principal components ((Table~\ref{tab:pca-variance}; 38.6\% explained by the top 10), while pooled skill embeddings concentrate variance sharply (76.2--80.0\% in the top 10, depending on pooling strategy). Pairwise cosine similarity confirms this collapse — text embeddings average 0.072 similarity between distinct samples, while pooled skill embeddings average 0.88--0.91, meaning skill representations for different career histories are nearly indistinguishable from one another (Figure~\ref{fig:pca-pooled}). Concatenated skill encoding partially mitigates this collapse (54.4\% variance in top 10 components; mean pairwise similarity 0.655) but remains substantially less discriminative than text (Figure~\ref{fig:pca-concat}).

This collapse compounds with redundancy: the mean cosine similarity between a career's text and skill embedding is 0.397, indicating that a substantial share of the skill embedding's content is already present in the text embedding. Finally, text embeddings are consistently closer to target occupation embeddings than skill embeddings (mean similarity 0.355 vs.\ 0.256), meaning text representations are simply better positioned in the latent space that ranking depends on. Together, these properties — low variance, redundancy with text, and weaker target alignment — explain why skill embeddings struggle to contribute discriminative signal regardless of integration strategy.

\subsection{Skill Genericness and Sample Difficulty}
\label{subsec:genericness}

A counterintuitive pattern emerges when stratifying performance by skill genericness, computed as the normalized inverse document frequency of predicted skills in DECORTE. Career paths with more generic skills (e.g., \textit{manage budgets}, \textit{manage staff}) achieve substantially higher accuracy across all models than those with specific skills (text-only Acc@1: 24.8\% for the most generic quartile vs.\ 11.8\% for the most specific), inverting the expected relationship in which specific skills should provide stronger discriminative signal (Table \ref{tab:genericness-quartiles}).

\begin{table}[t]
\centering
\caption{Performance by skill genericness quartile (DECORTE-based IDF). Higher genericness correlates with better performance across all models, reflecting dataset composition rather than skill utility.}
\label{tab:genericness-quartiles}
\begin{tabular}{lccc}
\toprule
Quartile & TEXT Acc@1 & SKILL Acc@1 & HYBRID Acc@1 \\
\midrule
Q1 (Specific) & 11.75\% & 10.20\% & 12.20\% \\
Q2            & 16.89\% & 11.11\% & 16.44\% \\
Q3            & 20.89\% & 14.89\% & 19.33\% \\
Q4 (Generic)  & 24.83\% & 23.50\% & 23.95\% \\
\bottomrule
\end{tabular}
\end{table}

This pattern is explained by dataset composition rather than skill utility: generic skills concentrate in well-represented occupation categories (e.g., Managers, Professionals) that dominate DECORTE, giving these career transitions more training examples and making them inherently easier to predict regardless of method. Correctly predicted samples have significantly lower genericness on average than incorrect ones across all models, including text-only (p \textless 0.001), confirming that genericness is a proxy for sample difficulty rather than a property of skill usefulness. A similar pattern holds for skill prediction confidence: text-only and skill-only accuracy improve in parallel from low- to high-confidence quartiles, indicating that confidence, too, reflects how easy a sample is overall rather than the quality of the skill signal specifically.

\begin{table}[t]
\centering
\caption{Skill enhancement effect by career trajectory type. Progression = target is Manager but last job is not; Regression = last job is Manager but target is not.}
\label{tab:trajectory-effects}
\begin{tabular}{lccc}
\toprule
Trajectory Type & Samples & MRR$_T$ & $\Delta$ MRR \\
\midrule
Lateral/Same-Level       & 1,291 & 0.298 & $-0.003$ \\
Progression (to Manager) & 263   & 0.216 & $-0.004$ \\
Regression (from Manager)& 248   & 0.150 & $+0.020$ \\
\bottomrule
\end{tabular}
\end{table}

\subsection{The Grounding Assumption Limitation}
\label{subsec:grounding}

A further limitation is structural rather than representational. Our skill prediction pipeline attributes skills at the occupation level: job titles are mapped to their most similar ESCO occupations, and skills are inherited from this prediction rather than extracted from the individual's actual described competencies. Two individuals holding the same job title therefore receive identical predicted skill sets, even if their real career histories differ. This grounding assumption caps the ceiling on complementary signal independent of model quality. Even a perfect occupation classifier would only recover the skills associated with an occupation in general, not those specific to the individual.

This limitation is consistent with the rank movement findings in Section~\ref{subsec:rank-movement}: skill enhancement recovers targets from very low ranks but rarely sharpens discrimination among similar occupations in the critical top-$k$ region. Related occupations in ESCO share substantial skill overlap by construction, so occupation-level skill attribution may identify the correct general neighborhood while lacking the granularity to distinguish between close alternatives — precisely the distinction that MRR and Recall@$k$ reward.

\subsection{Implications for Skill-Aware Recommender Design}
\label{subsec:implications}

\begin{table}
\caption{Skill enhancement effect by target ISCO-1 group (DECORTE test set). The four smallest ISCO-1 categories (Armed Forces, Agriculture, Craft \& Trades, Machine Operators; 57 samples combined) are omitted due to insufficient sample size for reliable estimates.}
\label{tab:isco-effects}
\begin{tabular}{lccc}
\toprule
Category & Count & MRR$_T$ & $\Delta$ MRR \\
\midrule
Professionals      & 617 & 0.309 & $+0.012$ \\
Managers           & 468 & 0.323 & $-0.014$ \\
Technicians        & 351 & 0.206 & $+0.004$ \\
Clerical Support   & 159 & 0.195 & $+0.005$ \\
Service \& Sales   & 137 & 0.194 & $-0.028$ \\
Elementary         & 13  & 0.119 & $+0.059$ \\
\bottomrule
\end{tabular}
\end{table}

These findings suggest two practical takeaways for practitioners building skill-aware recommender systems. First, explicit skill features should not be assumed to improve performance uniformly; their utility varies by occupational category (Table~\ref{tab:isco-effects}), helping Professionals the most ($+0.012$ MRR) while degrading Service \& Sales ($-0.028$). Checking redundancy at the category level, for instance via the score correlation introduced in Section~\ref{subsec:redundancy}, is a cheap diagnostic before investing in skill-prediction infrastructure for a given domain. Second, since skill enhancement helps primarily in metric-insensitive regions of the ranking while occasionally harming high-confidence predictions (Section~\ref{subsec:rank-movement}), selective integration that applies skill information only when textual confidence is low may be more productive than blanket fusion.

\subsection{Comparison with Prior Literature}
\label{subsec:prior-literature}

\citet{decorte2023careerpathpredictionusing} report a text-only baseline of MRR 0.247, improved to 0.274 by adding skill overlap. The text-only baseline we build on \citep{towards_more_realistic_2025}, which we reproduce at MRR 0.266, already exceeds their text-only result and approaches their skill-enhanced result without using any skill information. In addition, the two studies rely on different ESCO taxonomy versions (v1.1.2 vs.\ our v1.2.0), so the underlying occupation-skill associations are not directly comparable. Together, these differences suggest that skill contribution is highly sensitive to the strength of the text encoder and the taxonomy version used. Rather than being at odds with our findings, this pattern reinforces them: skill overlap appears to add value when the text representation leaves more headroom, and its contribution diminishes as the text encoder becomes more expressive, consistent with the redundancy we report in Section~\ref{subsec:redundancy}.

\section{Limitations}
\label{sec:limitations}

This study has three main limitations. First, we report single-run results without significance tests or variance estimates across random seeds, so small differences (e.g., 0.005--0.01 MRR) should be interpreted with caution. We mitigate this with an informal robustness argument: the same pattern, marginal and inconsistent gains from skill integration, holds across two datasets (DECORTE and KARRIEREWEGE+), multiple integration strategies (score fusion, skill pooling, skill concatenation), and both oracle and predicted skill conditions. This cross-setting consistency makes it unlikely that our central finding is an artifact of a single noisy run, though it does not substitute for formal statistical testing.

Second, our oracle condition is evaluated primarily in the score fusion setting (Section 4). In early exploratory runs, we also combined oracle skills with a neural architecture using ESCO-grounded job representations, but observed no improvement, consistent with the circularity discussed in Section~\ref{subsec:grounding}: skills derived from the occupation label add little when the job representation already encodes that occupation. This motivated our focus on free-text job representations, where occupational identity is less deterministic and skill information has more room to contribute.

Third, our experiments build on a single text-only baseline architecture, the MLP model of \citet{towards_more_realistic_2025}. It remains an open question whether our findings generalize to other CPP architectures.

\section{Conclusion}
\label{sec:conclusion}
We presented a systematic study of occupation-derived skill integration for career path prediction, finding that explicit skill integration provides only marginal and inconsistent gains over strong text-only baselines, even under oracle conditions with ground-truth skills. Our error analysis attributes this to high redundancy between textual and skill-based signals and severe representational collapse in skill embeddings. These findings suggest that practitioners should not assume skill features improve CPP by default, and that future work should explore selective integration strategies targeting cases of textual uncertainty, as well as skill signals extracted directly from individual career narratives rather than inherited from occupation labels.

\begin{acknowledgments}

We thank the reviewers for their insightful feedback. Elena Senger acknowledges financial support with funds provided by the German Federal Ministry for Economic Affairs and Climate Action due to an enactment of the German Bundestag under grant 46SKD127X (GENESIS).
\end{acknowledgments}

\section*{Declaration on Generative AI}
During the preparation of this work, the author(s) used Claude (Anthropic) in order to: assist with drafting and restructuring text, condense source material into prose, and perform grammar and language editing. After using this tool, the author(s) reviewed and edited all content as needed and take(s) full responsibility for the publication's content.

\bibliography{references}

\appendix

\end{document}